\documentclass[conference]{IEEEtran}

\usepackage[utf8]{inputenc}

\usepackage{csquotes}

\usepackage[numbers]{natbib}

\usepackage{graphicx}
\graphicspath{{pics/}}

\PassOptionsToPackage{hyphens}{url}\usepackage{hyperref}

\usepackage{array}

\usepackage{todonotes}
\usepackage{lipsum}

\usepackage{amsmath,amssymb,amsfonts}
\usepackage{algorithmic}
\usepackage{graphicx}
\usepackage{textcomp}
\usepackage{xcolor}
\usepackage{float}
\usepackage{caption}
\usepackage[many]{tcolorbox} 

\newtcolorbox{boxB}{
    fontupper = \color{black}, 
    boxrule = 1.5pt,
    colframe = black,
    rounded corners,
    arc = 5pt   
}

\def\BibTeX{{\rm B\kern-.05em{\sc i\kern-.025em b}\kern-.08em
    T\kern-.1667em\lower.7ex\hbox{E}\kern-.125emX}}
\begin{document}
\title{How to Sustainably Monitor ML-Enabled Systems? Accuracy and Energy Efficiency Tradeoffs in Concept Drift  Detection}

\author{
   \IEEEauthorblockN{
       Rafiullah Omar\IEEEauthorrefmark{1}, Justus Bogner\IEEEauthorrefmark{2}, Joran Leest\IEEEauthorrefmark{2}, Vincenzo Stoico\IEEEauthorrefmark{2}, Patricia Lago\IEEEauthorrefmark{2}, Henry Muccini\IEEEauthorrefmark{1}
   }
   \IEEEauthorblockA{
       \IEEEauthorrefmark{1}FrAmeLab, University of L’Aquila, L'Aquila, Italy,\\ rafiullah.omar@graduate.univaq.it, henry.muccini@univaq.it
   }
    \IEEEauthorblockA{
       \IEEEauthorrefmark{2}Vrije Universiteit Amsterdam, Amsterdam, The Netherlands,\\ j.bogner@vu.nl, j.g.leest@vu.nl, v.stoico@vu.nl, p.lago@vu.nl
   }
}

\maketitle


\begin{abstract}
ML-enabled systems that are deployed in a production environment typically suffer from decaying model prediction quality through \textit{concept drift}, i.e., a gradual change in the statistical characteristics of a certain real-world domain.
To combat this, a simple solution is to periodically retrain ML models, which unfortunately can consume a lot of energy.
One recommended tactic to improve energy efficiency is therefore to systematically monitor the level of concept drift and only retrain when it becomes unavoidable.
Different methods are available to do this, but we know very little about their concrete impact on the tradeoff between accuracy and energy efficiency, as these methods also consume energy themselves.

To address this, we therefore conducted a controlled experiment to study the accuracy vs. energy efficiency tradeoff of seven common methods for concept drift detection.
We used five synthetic datasets, each in a version with abrupt and one with gradual drift, and trained six different ML models as base classifiers.
Based on a full factorial design, we tested 420 combinations (7 drift detectors $\times$ 5 datasets $\times$ 2 types of drift $\times$ 6 base classifiers) and compared energy consumption and drift detection accuracy.

Our results indicate that there are three types of detectors: a) detectors that sacrifice energy efficiency for detection accuracy (KSWIN), b) balanced detectors that consume low to medium energy with good accuracy (HDDM\_W, ADWIN), and c) detectors that consume very little energy but are unusable in practice due to very poor accuracy (HDDM\_A, PageHinkley, DDM, EDDM).
By providing rich evidence for this energy efficiency tactic, our findings support ML practitioners in choosing the best suited method of concept drift detection for their ML-enabled systems.
\end{abstract}



\section{Introduction}
More and more industry domains employ machine learning (ML)~\cite{Jordan2015} for advanced prediction functionality.
Once an ML model with sufficient prediction quality has been trained, it is integrated into an ML component that is deployed as part of an ML-enabled system~\cite{Martinez-Fernandez2022}.
However, the successful development and especially the long-term operation of ML-enabled systems requires considerable expertise due to many challenges~\cite{Nahar2023,Paleyes2023}.
In rapidly evolving domains, one particular problem is the dynamic nature of real-world data.
This phenomenon is known as \textit{concept drift}, i.e., a decline in predictive accuracy over time due to changing statistical relationships in the underlying data distribution of the real-world phenomenon~\cite{Gama2014}.
For most use cases, this means that deployed ML models will unavoidably decay over time without periodic retraining based on newly collected data.
To identify when this happens in a production system, sophisticated monitoring capabilities are required, which requires the development of robust \textit{drift detectors}.
By now, a wide variety of such techniques have been proposed~\cite{Lu2018,Barros2018} and implemented for popular ML frameworks, such as \texttt{scikit-multiflow}.\footnote{\url{https://scikit-multiflow.github.io}}

The primary quality concern for drift detectors is obviously their ability to quickly and correctly identify drift, which in turn is the foundation for improving the accuracy of the monitored model through retraining.
However, there is evidence that drift detectors have a negative impact on the energy consumed by ML-enabled systems \cite{Garcia-Martin2019}, which already require significant amounts of energy that lead to carbon dioxide emissions.
Although these systems can also be used to improve environmental sustainability, e.g., ML-enabled systems that support the identification of water contaminants and their recycling\footnote{\url{https://www.cbsnews.com/news/water-reuse-recycling-toilet-to-tap-yuck-factor/}}, the emissions generated by their development and operation are a rising concern~\cite{Schwartz2020, jay2024prioritize, freitag2021real}.
For example, the training phases of LLaMA~\cite{Touvron_2023} and Code LLaMA~\cite{Roziere_2023} released a total of 1,015 tons of carbon emission (tCO2eq) and 63.5 tCO2eq respectively.

Since (re-)training a complex model (called the \textit{base classifier} in our context) can consume substantial amounts of energy~\cite{Strubell2019}, simple periodic retraining is often not a solution that favors energy efficiency.
Instead, relying on \textit{informed adaptation} through monitoring and only retraining when necessary is a recommended tactic in this space~\cite{Järvenpää2023}.
While various researchers have analyzed the accuracy of drift detectors~\cite{Barros2018,Choudhary2022}, there is little evidence and no fine-grained analysis examining the impact of different drift detectors on the energy consumed by ML-enabled systems at runtime.
Drift detection can increase the energy consumed by the system in two main ways: a) the energy consumed by the detectors themselves, and b) the energy consumed through model retraining once a detector triggers an alarm.
A single instance of the first is much smaller than the second, but this energy is also continuously expended during system operation, and potentially also for multiple ML models.
For the second, the decision to retrain an ML model leads to the consumption of a substantial amount of energy, which means that the accuracy of a detector also indirectly impacts energy efficiency at the system level.
To make an informed decision about which drift detector to use under which conditions, it is important to look at these two quality attributes holistically.

In this experiment, we focus on the tradeoff between energy efficiency vs. accuracy while using various concept drift detectors for retraining decisions and examine their performance across synthetic datasets.
The exploration of energy consumption holds paramount importance, as it directly influences the practical viability of these detectors in real-world applications.
By shedding light on the energy dynamics of concept drift detection, this study aims to contribute valuable insights to the optimization of ML-enabled systems, ensuring their effectiveness and environmental sustainability in dynamic, data-evolving scenarios.
ML researchers and practitioners can use our results to guide their selection of drift detectors.
Our study answers the following research questions:

\textbf{RQ1:} How does the tradeoff between accuracy and energy efficiency manifest in different concept drift detectors?

Ideally, a drift detector would exhibit very good accuracy and energy efficiency at the same time.
However, it is also possible that some detectors sacrifice energy efficiency for accuracy or vice versa.
We want to analyze this relationship for different detectors.

\textbf{RQ2:} How does the type of dataset influence this tradeoff?

It is possible that different types of datasets change how this tradeoff manifests, e.g., for abrupt drift or gradually introduced drift.
Existing research has shown differences in accuracy based on different drift datasets~\cite{Barros2018}, so it is natural to assume that differences do exist for energy consumption as well.

\textbf{RQ3:} How does the type of base classifier influence this tradeoff?

Additionally, it is also possible that the type of the monitored ML model (the base classifier) influences the results, which previous research showed for accuracy~\cite{Barros2018}.
Since we also holistically include the energy consumption of retraining the base classifier, this tradeoff might manifest differently for different types of ML models.

In the following sections, we first explain the necessary background for our study and present related work in the area.
We then describe our detailed experiment design, provide the results, and discuss their implications.
Lastly, we comment on threats to the validity of our study and present a conclusion.

\section{Background and Related Work}

\subsection{Concept Drift in ML}
ML-enabled systems are unique software systems with behavior that is heavily influenced by their application context~\cite{huyen2022}.
This context dependence poses a challenge in maintaining ML model prediction quality, especially when faced with non-stationary contexts that affect the data distribution.
We can identify two primary causes of concept drift affecting model accuracy~\cite{huyen2022, shankar2021towards}.
The first cause is external events that arise in the environment of the ML-enabled system.
Examples of these events include marketing campaigns targeting specific demographics, shifts in user behavior, and changes to the user interface.
The second cause of concept drift in ML-enabled systems stems from data integrity issues~\cite{shankar2022operationalizing}.
These arise from continuous updates in the data pipeline, a common aspect of the dynamic workflow in ML engineering.
Such updates can accidentally introduce errors or bugs, such as swapping columns or representing a column with the wrong unit of measurement, e.g., using miles instead of kilometers. 

\begin{figure}[ht]
    \centering
    \includegraphics[width=\linewidth]{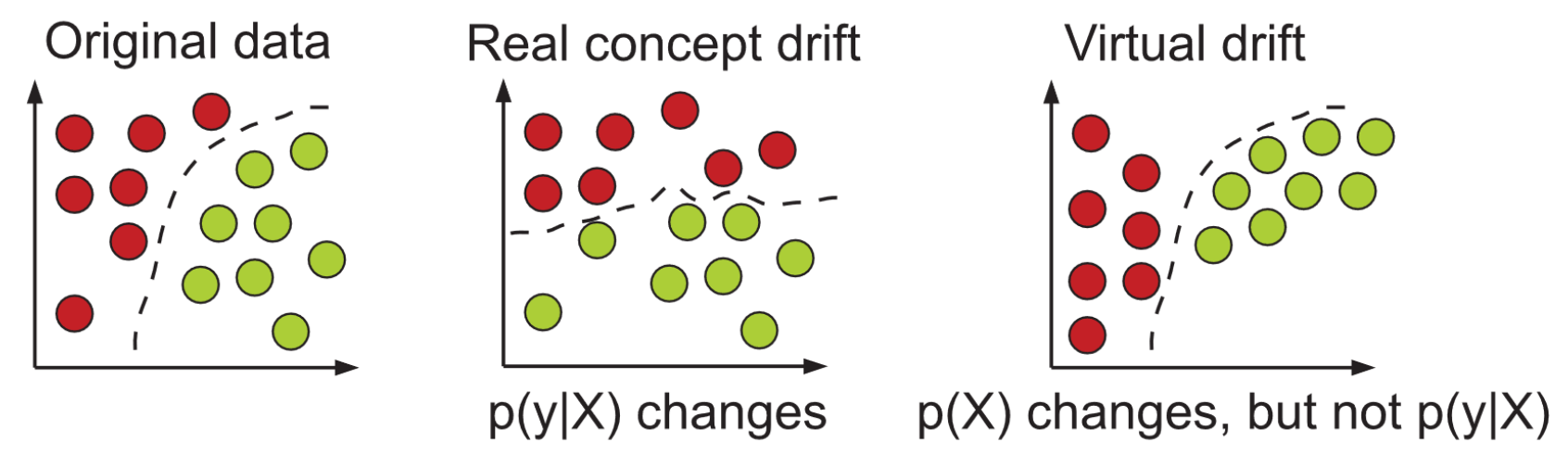}
    \caption{Difference between concept drift (middle) and virtual drift (right), taken from \citet{Gama2014}}
    \label{fig:concept-drift-vs-virtual-drift}
\end{figure}

The impact of distribution changes on model accuracy, whether due to external events or data integrity issues, varies based on the severity and type of drift.
We differentiate between \textit{virtual drift} (also known as feature drift or covariate shift), which is a change in the input distribution $P(X)$, and \textit{real concept drift} (or model drift), a change in the conditional probability distribution of the target variable for a given input $P(Y|X)$~\cite{Gama2014}.
Virtual drift typically has a benign effect, impacting prediction quality only when new observations fall into areas of the training distribution that are uncertain or underrepresented.
In contrast, real concept drift always leads to a deterioration in model accuracy, as the underlying relationship is no longer captured by the learned mapping function.
This difference is visualized in Fig.~\ref{fig:concept-drift-vs-virtual-drift}.

\begin{figure}[ht]
    \centering
    \includegraphics[width=\linewidth]{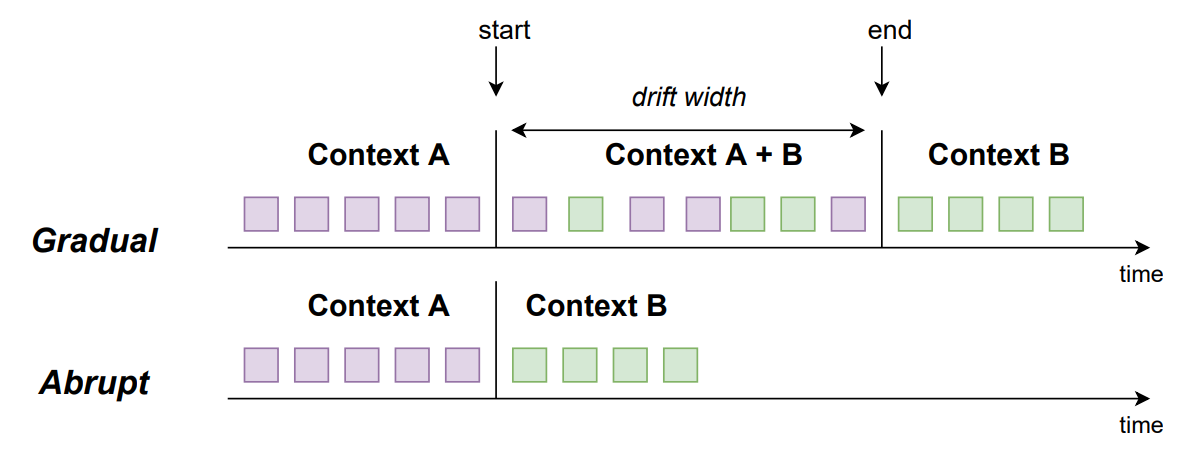}
    \caption{Gradual concept drift (top) vs. abrupt concept drift (bottom), taken from \citet{Poenaru-Olaru2022}}
    \label{fig:gradual-vs-abrupt-drift}
\end{figure}

The nature of the cause of concept drift influences its characteristics, such as abruptness or gradualness, severity, recurrence, and duration~\cite{webb2016characterizing}.
These factors are important to consider when setting up monitoring systems and selecting concept drift detection algorithms.
For example, data integrity issues often cause abrupt changes, leading to immediate shifts in the data distribution~\cite{shankar2021towards}.
External events, in contrast, tend to result in more gradual changes, manifesting as a slow but constant evolution of the data distribution.
Gradual drift therefore has a start and an end, after which the change becomes permanent.
The distance between start and end is called \textit{drift width} (see Fig.~\ref{fig:gradual-vs-abrupt-drift}).

Concept drift has been a research topic in the data mining community for decades, leading to the development of various detectors and mitigation methods~\cite{Lu2018}.
Traditionally, these tools are designed under the assumption of immediate label feedback, where the label for a prediction is available shortly after the prediction is made.
Under this assumption, it is not necessary to distinguish between types of drift.
Instead, model accuracy serves as a general indicator of drift, enabling the use of a single drift detector that analyzes model errors to identify distribution shifts.
In this work, we focus on settings with immediate label feedback, allowing drift detectors to identify any type of drift by directly comparing model predictions to the true labels.

\subsection{Energy Consumption of ML-Enabled Systems}
In recent years, ML models and systems that integrate them have been more and more associated with large energy and carbon footprints.
For example, training a common natural language processing (NLP) model based on transformers can lead to greenhouse gas emissions that are similar to those produced by several cars throughout their lifetime~\cite{Strubell2019}.
As a result, \citet{Schwartz2020} coined the distinction between \textit{Green AI} and \textit{Red AI}.
Historically, AI development was primarily aimed at achieving high prediction quality without caring too much about energy efficiency (Red AI).
Green AI development, on the other hand, is conscious about the potential environmental impact of AI and tries to minimize energy consumption and carbon emissions without substantially impacting accuracy.

While accuracy and energy efficiency are still often regarded as a tradeoff~\cite{Brownlee2021}, studies have shown that there are many techniques to substantially reduce energy consumption with a negligible decrease in accuracy~\cite{Verdecchia2022,DelRey2023,Wei2023}.
Today, Green AI is an active field of research, with a recent review from ~\citet{Verdecchia2023} identifying 98 primary studies on the topic.
As the initial development of ML models is commonly considered the most energy-intensive phase within the life cycle of ML-enabled systems~\cite{Kaack2022}, making ML training more energy-efficient receives a lot of attention.
However, other phases like the operation of ML-enabled systems also hold much potential for reducing their environmental footprint.
In this study, we focus on the continuous operation phase of such systems.

\subsection{Related Work}
In the exploration of software energy consumption, various studies have investigated the impact of different programming languages~\cite{8595213}, data structures~\cite{8816747}, and frameworks~\cite{CALERO2021100603, 10.1145/3510003.3510221}.
With the advancing progress in AI and the inherently energy-intensive nature of AI systems, there is a growing demand for energy in both training and maintaining these systems~\cite{schwartz2019green}.
Recent research has honed in on the energy consumption of ML systems, aiming to mitigate the energy demands of ML-enabled systems~\cite{Verdecchia2023, Järvenpää2023}.
Different studies have concentrated on energy consumption of various phases of ML system development.
For instance, Verdecchia et al.~\cite{Verdecchia2022} focused on data-centric approaches to minimize energy consumption during training.
In the context of evolving data that influences and degrades ML accuracy, concept drift requires model retraining~\cite{Poenaru-Olaru2023}.
Strategies addressing concept drift detection and management have been developed~\cite{Barros2018}, and comparative studies, such as the one by Goncalves et al.~\cite{GONCALVES20148144}, have assessed the accuracy of different drift detectors.
In the realm of drift detection, studies, including those by Barros et al.~\cite{Barros2018}, have scrutinized various methods using different classifiers and datasets.
However, this analysis lacks an essential metric: energy consumption.
We address this gap by conducting a comprehensive comparison of different drift detection methods.
While accuracy is considered, our primary focus centers on analyzing the tradeoffs between accuracy and energy, shedding light on the energy efficiency of these diverse detection methods.
To the best of our knowledge, no other study has so far considered this tradeoff in detail.

\section{Experiment Design}
We designed a controlled experiment to answer our research questions~\cite{Wohlin2012}.
In this section, we describe the important details associated with this research method, namely our experiment objects (concept drift detectors, datasets, and base classifiers), experiment variables, experiment execution, and data analysis.

\subsection{Concept Drift Detectors}
Many different methods for concept drift detection have been studied and implemented for popular ML frameworks.
Since the chosen framework may also affect energy consumption, it is important to focus on a single framework.
For an accurate comparison, we therefore selected all seven concept drift detectors implemented in \texttt{scikit-multiflow}\footnote{\url{https://scikit-multiflow.github.io}}: ADWIN, DDM, EDDM, HDDM-A, HDDM-W, KSWIN, and PageHinkley.
These methods are usable within the same popular ML framework, \texttt{scikit-learn}\footnote{\url{https://scikit-learn.org}}, and are also among the most used detectors in studies about concept drift~\cite{Barros2018}.
They also employ reasonably different strategies that could affect their energy consumption.
For a fair comparison, we implemented each detector with its default configuration.
We briefly describe each detector below.
For more details, please refer to the official documentation of each detector.\footnote{\url{https://scikit-multiflow.readthedocs.io/en/stable/api/api.html\#module-skmultiflow.drift_detection}}

\begin{itemize}
    \item ADWIN (Adaptive Windowing) dynamically adjusts its window size based on statistical measures. It monitors average and variance for incoming data to detect changes for both gradual and abrupt concept drift.
    \item DDM (Drift Detection Method) is a statistical approach based on the standard deviation of performance measures to identify changes. It triggers an alert when the deviation exceeds a predefined threshold.
    \item EDDM (Exponentially Weighted Moving Detection Method) extends DDM by assigning exponentially decreasing weights to older instances. It gives priority to recent data while being robust against noise.
    \item HDDM\_A (Hoeffding's Drift Detection Method - Adaptive) is an adaptive version of Hoeffding's method, using the Hoeffding bound to detect data distribution changes and adjusting sensitivity based on observed variance.
    \item HDDM\_W (Hoeffding's Drift Detection Method - Window) is a window-based adaptation of Hoeffding's method. It maintains a fixed-sized window to monitor mean changes using the Hoeffding bound.
    \item KSWIN (Kernelized Sliding Window INterval) applies a kernel function to data using a sliding window. It monitors the estimated kernel density for significant changes.
    \item PageHinkley computes a cumulative sum of differences between expected and observed values. It triggers an alert when the cumulative sum exceeds a predefined threshold.
\end{itemize}

\subsection{Datasets}
To connect our experiment to previous work, we selected five popular synthetic datasets available in the Harvard Dataverse\footnote{\url{https://dataverse.harvard.edu}}, a public repository of research data.
This collection of drift datasets was curated by \citet{Lobo2020} via stream generators.
In January 2024, it had accumulated over 7,600 downloads.
The datasets have been used in similar studies~\cite{Choudhary2022} and feature both gradual and abrupt concept drifts.
To incorporate some diversity into our experiment, we selected five different datasets from the collection:

\begin{itemize}
    \item \texttt{sine}: two numeric attributes (X1 and X2), where the class is assigned based on the sine function
    \item \texttt{stagger}: three nominal attributes (X1, X2 and X3)
    \item \texttt{mixed}: two numeric attributes ($x_1$ and $x_2$) and two boolean attributes ($x_3$ and $x_4$)
    \item \texttt{sea}: three numeric attributes ($x_1$, $x_2$, and $x_3$)
    \item \texttt{RT}: two numeric attributes ($x_1$ and $x_2$)
\end{itemize}

Each of the five datasets is available in two versions: one with abrupt drift and one with gradual drift.
Notably, the dataset creator introduced concept drift into each dataset in three locations.
At each location, i.e., at each occurrence of concept drift, the classification is reversed, or in other words, the conditional probability $P(Y|X)$ changes.
Each abrupt dataset consists of 40,000 instances and gradual dataset 41,000 instances.
The drift is positioned at specific intervals: 10,000, 20,000, and 30,000 instances for abrupt drift datasets, and 9,500, 20,000, and 30,500 instances for gradual drift datasets.
The width for the gradual drift is 1,000, i.e., the drift starts around time step 9,500 and increases for 1,000 instances, covering the range from 9,500 to 10,500.
Similarly, another gradual drift occurs around time step 20,000 and covers instances from 20,000 to 21,000.
The third gradual drift starts around time step 30,500 and covers instances from 30,500 to 31,500. 
In contrast to real-world datasets where the precise start of concept drift cannot be objectively determined, the artificial placement of drifts facilitates a controlled environment for evaluating concept drift detectors, simplifying the comparison of results.
This pre-defined drift position is particularly advantageous, offering a clear and consistent basis for assessing the tradeoff between accuracy and energy consumption.

\subsection{Base Classifiers}
For our drift detector comparison, we also needed ML models as base classifiers to predict the class of the incoming data streams, which is compared to the correct class.
In similar studies, two ML algorithms are commonly used as base classifiers: Naive Bayes and Decision Tree.
Since these classifiers differ in terms of energy efficiency, we also want to understand their effects on the drift detector and the overall energy consumption, which includes the retraining of models.
Therefore, in addition to Naive Bayes and Decision Tree, we selected five more classifiers that have been studied in previous work~\cite{Verdecchia2022} for their energy efficiency during the training phase.
We want to investigate whether previous findings remain valid for these classifiers when dealing with concept drift.
The six included classifiers are Support Vector Machine (SVM), Hoeffding Tree, k-nearest neighbors (KNN), Random Forest, AdaBoost, Bagging Classifier.

This selection provides a decent variety for analyzing the energy consumption in this context.
For each classifier, we used the \texttt{scikit-learn} implementation with the default configuration.
Please refer to the official documentation for additional details about each classifier.\footnote{\url{https://scikit-learn.org/stable/supervised_learning.html}}
Note that we consciously focused on more traditional ML algorithms and excluded deep learning.
While different neural network architectures would also be interesting candidates for such analyses, they are outside the scope of this study.

\subsection{Experiment Variables}
We focus on the comparison of two \textbf{dependent variables}:

\begin{itemize}
    \item \textit{Energy consumption} measured in Joule (J); this includes the energy consumed by the drift detector plus the retraining of base classifiers when drift was detected.
    \item \textit{Detection accuracy} for each drift detector; if a detector correctly identified drift within the window (start of new concept until end of concept), we documented it as a \textit{true alarm}. Warning about drift outside of the window was documented as a \textit{false alarm}. We calculated the \textit{true alarm percentage}. For the true alarms, we additionally calculate how quickly the drift was detected, i.e., the distance between the start of the drift and the instance for which drift was detected.
\end{itemize}

Based on our experiment objects, we had several \textbf{independent variables} that we manipulated to see how they influence the dependent variables:
\begin{itemize}
    \item \textit{Drift detector}: ADWIN, DDM, EDDM, HDDM\_A, HDDM\_W, KSWIN, PageHinkley
    \item \textit{Dataset}: sine, mixed, RT, sea, stagger
    \item \textit{Type of drift}: abrupt, gradual
    \item \textit{Base classifier}: Naive Bayes, SVM, Decision Tree, KNN, Random Forest, AdaBoost, Bagging Classifier
\end{itemize}
The different drift detectors were the primary independent variables (\textit{treatments}) that we manipulated for RQ1.
The synthetic datasets and the types of drift were manipulated for RQ2, with the base classifier manipulation answering RQ3.
Since this study was exploratory in nature, we did not have any preconceived hypotheses to confirm.

\subsection{Experiment Execution}
Based on the different independent variables, we opted for a \textit{full factorial design}~\cite{Wohlin2012}, meaning we had an experiment space of 7 drift detectors $\times$ 5 datasets $\times$ 2 types of drift $\times$ 6 base classifiers (420 combinations).
For each combination, we followed these steps, which constitute one \textit{iteration} in our experiment:

\begin{enumerate}
    \item Train the initial base classifier on the first 8.5k instances of the dataset.
    \item Record energy consumption and accuracy.
    \item Predict labels for the remaining instances of the dataset.
    \item Individually pass each predicted and true label to the drift detector.
    \item If drift is detected, stop detection and record drift position and consumed energy for drift detection.
    \item Calculate base classifier accuracy on data from 10k to the drift position.
    \item Retrain the base classifier on the dataset from 0 to the drift position and record energy and accuracy.
\end{enumerate}

Before the initiation of the experiment, a warm-up function was executed to stabilize the system and optimize its performance before data collection began.
To minimize the effects of non-determinism and potential fluctuations in the experiment infrastructure, e.g., other processes briefly consuming more energy than usual, each iteration was repeated 10 times.
Additionally, 5-seconds sleep interval was incorporated between each iteration to ensure that the infrastructure conditions could return closer to the initial state.
These measures collectively contributed to the reliability of our experimental framework, allowing for a thorough and dependable assessment of the energy consumption of concept drift detectors in machine learning.

For the experiment instrumentation, we wrote several scripts in Python, the de facto standard language for data and ML practitioners, which automated the complete process described above.
To estimate the energy consumption, we used the CodeCarbon software package.\footnote{\url{https://codecarbon.io}}
While CodeCarbon is not as precise as hardware-based measurements, it has been shown to underreport the actual energy consumption only by a few percentage points~\cite{Xu2023}, with very strong correlation between both measurements ($\rho=0.94$).
It is therefore accepted practice to use software-based estimation tools like CodeCarbon when the focus of the study is to compare differences in energy consumption between several variants.
The complete reproducible code is publicly available in our artifact repository.\footnote{\url{https://doi.org/10.5281/zenodo.10613150}}
We ran the study on dedicated experiment infrastructure for software energy experiments located at our university.
The infrastructure contains a server equipped with 36 TB HDD, 384 GB RAM, and an Intel Xeon CPU including 16 cores with hyperthreading running at 2.1GHz (i.e., 32 vCPUs).
Access to the server was restricted during experiment execution to prevent unwanted additional load.

\subsection{Data Analysis}
As an initial step, we assessed the normality of the distribution of results with the Shapiro-Wilk test~\cite{Shapiro1965}.
The obtained result indicates that the data exhibits non-normal distribution characteristics, as evidenced by a p-value of $<$= 5.6e-45.
For the purpose of comparing and identifying significant differences in the energy consumption among various drift detectors, we therefore employed the Mann-Whitney U test~\cite{Mann1947}, which can handle such distributions.
To compare the magnitude of effects, we computed the percentage difference between the mean energy consumption values of distinct drift detectors.
Furthermore, Cohen's d values were computed for a more comprehensive understanding of the effect size~\cite{Cohen1988}.
In exploring the influence of drift types, we conducted analogous tests to those used for comparing drift detectors.
Since we performed many pair-wise hypothesis tests, the Holm-Bonferroni correction was used to combat the multiple comparison problem~\cite{Shaffer1995}.
To investigate the impact of base classifiers on both the energy consumption of drift detectors and their accuracy, we applied the Spearman correlation test~\cite{Myers2002}.

\section{Results}
In this section, we present the experiment results according to the research questions.
\subsection{RQ1: How does the tradeoff between accuracy and energy efficiency manifest in different concept drift detectors?}
\begin{figure}[ht]
    \centering
    \includegraphics[width=1.0\linewidth]{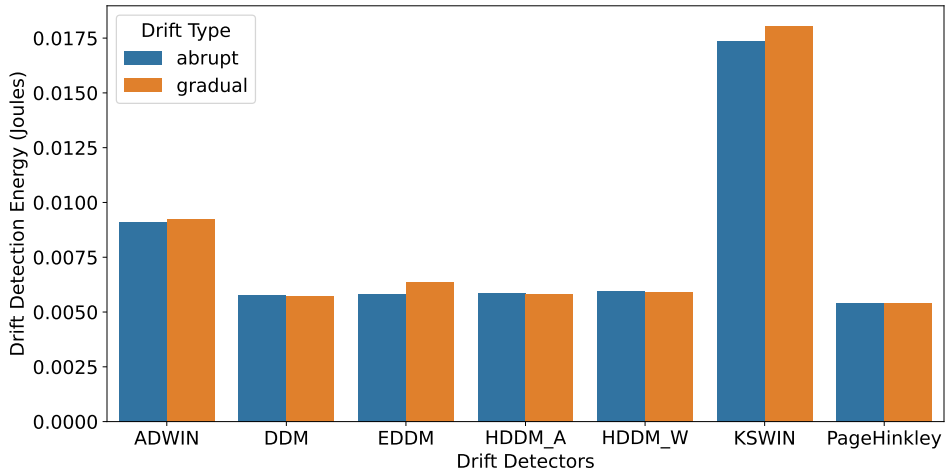} 
    \caption{Energy consumption of different drift detectors and drift types}
    \label{fig:energy-consumption-gradual-abrupt}
\end{figure}

We delved into the energy consumption of various concept drift detectors, and the results depicted in Fig.~\ref{fig:energy-consumption-gradual-abrupt} revealed several variations among them.
Notably, the most energy-intensive method for both abrupt and gradual drifts was found to be KSWIN.
Following closely as the second highest consumer of energy was ADWIN, while the remaining detectors displayed varying levels of energy consumption, albeit with less pronounced differences.

\begin{figure}[ht]
    \centering
    \includegraphics[width=0.9\linewidth]{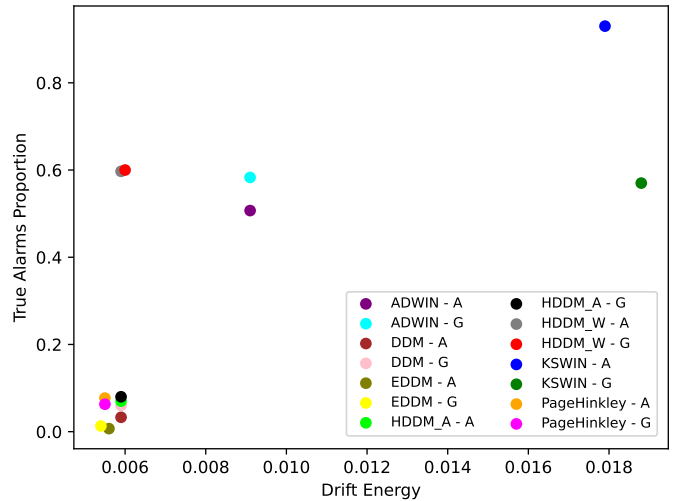} 
    \caption{True alarm percentage vs. energy consumption}
    \label{fig:energy-consumption-true-alarm}
\end{figure}

\begin{figure}[ht]
    \centering
    \includegraphics[width=0.9\linewidth]{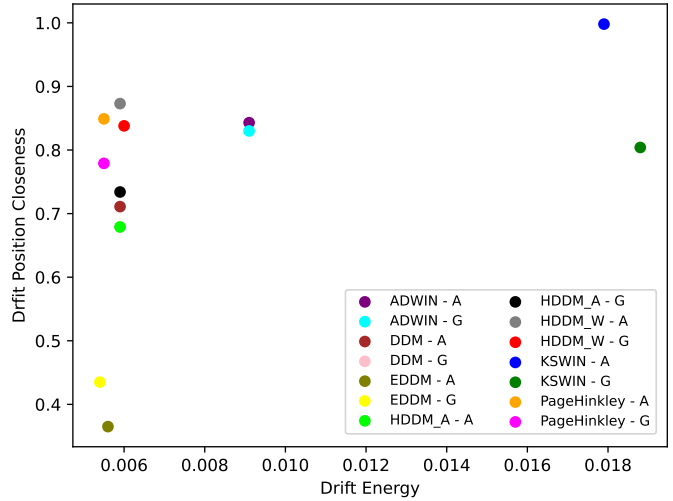} 
    \caption{Drift detection closeness for true alarms vs. energy consumption}
    \label{fig:energy-consumption-closeness}
\end{figure}

\begin{table*}[!ht]
    \centering
    \caption{Pair-wise Mann-Whitney U tests for energy consumption of detectors (Holm-Bonferroni adjusted p-values, $\alpha=0.05$, significant p-values in bold)}
    \begin{tabular}{llrrrrr}
        \textbf{Detector 1} & \textbf{Detector 2} & \textbf{p-value} & \textbf{Detector 1 Mean} & \textbf{Detector 2 Mean} & \textbf{Difference (\%)} & \textbf{Cohen's d} \\
        \hline
        \hline
        KSWIN & PageHinkley & \textbf{0.00013} & 0.0181 & 0.0055 & 69.8 & 7.53  \\
        KSWIN & HDDM\_A & \textbf{0.00015} & 0.0181 & 0.0058 & 67.7 & 7.20  \\
        KSWIN & DDM & \textbf{0.00011} & 0.0181 & 0.0059 & 67.6 & 7.24  \\
        KSWIN & EDDM & \textbf{0.00012} & 0.0181 & 0.0060 & 67.0 & 6.82  \\
        KSWIN & HDDM\_W & \textbf{0.00013} & 0.0181 & 0.0060 & 67.0 & 7.16  \\
        KSWIN & ADWIN & \textbf{0.00014} & 0.0181 & 0.0091 & 49.4 & 4.97  \\
        ADWIN & PageHinkley & \textbf{0.00034} & 0.0091 & 0.0055 & 40.2 & 6.16  \\
        ADWIN & HDDM\_A & \textbf{0.00022} & 0.0091 & 0.0058 & 36.2 & 5.15  \\
        ADWIN & DDM & \textbf{0.00024} & 0.0091 & 0.0059 & 35.9 & 5.20  \\
        ADWIN & HDDM\_W & \textbf{0.00030} & 0.0091 & 0.0060 & 34.7 & 5.07  \\
        ADWIN & EDDM & \textbf{0.00026} & 0.0091 & 0.0060 & 34.7 & 3.98  \\
        PageHinkley & EDDM & \textbf{0.00048} & 0.0055 & 0.0060 & 08.5 & 1.25  \\
        PageHinkley & HDDM\_W & \textbf{0.00060} & 0.0055 & 0.0060 & 08.5 & 1.47  \\
        PageHinkley & DDM & \textbf{0.00040} & 0.0055 & 0.0059 & 06.7 & 0.90  \\
        HDDM\_A & PageHinkley & \textbf{0.00020} & 0.0058 & 0.0055 & 06.3 & 1.16  \\
        HDDM\_A & EDDM & \textbf{0.00017} & 0.0058 & 0.0060 & 02.3 & 0.44  \\
        HDDM\_A & HDDM\_W & \textbf{0.00018} & 0.0058 & 0.0060 & 02.3 & 0.23  \\
        EDDM & DDM & \textbf{0.00238} & 0.0060 & 0.0059 & 01.9 & 0.57  \\
        HDDM\_W & DDM & \textbf{0.00079} & 0.0060 & 0.0059 & 01.9 & 0.41  \\
        HDDM\_A & DDM & \textbf{0.00016} & 0.0058 & 0.0059 & 00.4 & 0.19  \\
        HDDM\_W & EDDM & \textbf{0.00119} & 0.0060 & 0.0060 & 0.00 & 0.28  \\
        \hline
        \hline
    \end{tabular}
    \label{tab:energy-drift-detectors}
\end{table*}

These findings were further substantiated through statistical analysis, employing the Mann-Whitney U test with p-values adjusted using the Holm-Bonferroni method.
The pair-wise tests of the differences in overall energy consumption of all detectors resulted in significant differences for each comparison.
To quantify the extent of the difference between each pair of methods, we calculated the percentage differences and Cohen's d values.
The maximum disparity was noted between KSWIN and PageHinkley (69.78\%), implying that employing KSWIN over PageHinkley consumed nearly 70\% more energy, with very large effect size of $d=7.53$.
Similarly, KSWIN consumed 49.44\% more energy than ADWIN.
On the other end of the spectrum, the minimum difference was observed between HDDM\_W and EDDM, totaling a mere 0.02\%.  DDM, DDM\_A, and DDM\_W, along with EDDM, exhibit relatively equal energy consumption, with differences below 2.3\%.
Therefore, the effect sizes are small between DDM, DDM\_A, DDM\_W, and EDDM, as their Cohen's d values are below 0.5.
Detailed information regarding the test and mean differences between methods is presented in Table~\ref{tab:energy-drift-detectors}.

\begin{table*}[!ht]
    \centering
    \caption{Accuracy of drift detectors for abrupt and gradual drift (ordered by true alarm percentage)}
    \begin{tabular}{llrrrrrrr}
        \textbf{Drift Detector} & \textbf{Drift Type}  & \textbf{\# of True Alarms} & \textbf{\# of False Alarms} & \textbf{\# of Missed Alarms} & \textbf{True Alarm \%} & \textbf{Mean Detection Closeness} \\
        \hline
        \hline
        KSWIN & abrupt  & 279 & 21 & 0 & 93.0  & 0.998 \\
        HDDM\_W & gradual  & 180 & 120 & 0 & 60.0  & 0.838 \\
        HDDM\_W & abrupt  & 179 & 121 & 0 & 59.7  & 0.873 \\
        ADWIN & gradual  & 175 & 118 & 7 & 58.3 & 0.830 \\
        KSWIN & gradual  & 171 & 129 & 0 & 57.0  & 0.804 \\
        ADWIN & abrupt  & 152 & 136 & 12 & 50.7  & 0.843 \\
        HDDM\_A & gradual  & 24 & 243 & 33 & 8.0  & 0.734 \\
        PageHinkley & abrupt  & 23 & 259 & 18 & 7.7  & 0.849 \\
        HDDM\_A & abrupt  & 21 & 240 & 39 & 7.0 & 0.679 \\
        PageHinkley & gradual & 19 & 252 & 29 & 6.3  & 0.779 \\
        DDM & gradual  & 18 & 240 & 42 & 6.0  & 0.679 \\
        DDM & abrupt  & 10 & 252 & 38 & 3.3  & 0.711 \\
        EDDM & gradual  & 4 & 223 & 73 & 1.3  & 0.435 \\
        EDDM & abrupt  & 2 & 242 & 56 & 0.7 & 0.365 \\
        \hline
        \hline
    \end{tabular}
    \label{tab:accuracy-drift-detectors}
\end{table*}

Throughout these experiments, drift positions were recorded, enabling the assessment of the accuracy of alarms in relation to the actual drift position.
This involved calculating the percentages of true alarms, number of false alarms, and missed alarms.
Additionally, the relative closeness of the detected drift to the actual drift position was quantified, with values ranging between 0 and 1, i.e., the higher the closeness value, the smaller the distance between the detected drift position and the actual start drift position.
Values close to 1 indicate the detector identified the drift extremely quickly, while values close to 0 indicate very late detection.
These results are displayed in Table~\ref{tab:drift-type-energy-comparison}.
In terms of accuracy, KSWIN emerged as the top-performing method in abrupt drift scenarios, with an impressive 93\% true alarm rate and 0.998 closeness.
For gradual drift, however, its performance was notably worse (57\%, 0.804).
In these scenarios, HDDM\_W took first place (60\%, 0.838).
Its similarly good performance for abrupt drift (59.7\%, 0.873) positioned it as a balanced and top-performing method.
The fairly balanced ADWIN came in third.
It performed slightly better for gradual drift scenarios (58.3\% vs. 50.7\%) than for abrupt ones (50.7\%, 0.843).
Interestingly, all other four methods (HDDM\_A, PageHinkley, DDM, EDDM) can be considered useless for production systems based on our results.
Even though the detection closeness for true alarms was decent for some of these methods, their overall true alarm rate was below 10\% in all cases.
The least effective method was EDDM.
Considering missed alarms, we observe that the best-performing methods in drift detection closeness and true alarms percentage also excel in minimizing missed alarms.
Importantly, the results highlighted the absence of a singular method excelling across all compared metrics.
It is crucial to acknowledge that this comparison is generalized and does not account for specific datasets or base classifiers used.
The nuanced impact of drift type and base classifier is scrutinized in greater detail in RQ2 and RQ3.

\begin{boxB}
\textbf{Answer to RQ1:}
KSWIN trades off energy efficiency for very good detection accuracy.
It consumes by far the most energy.
ADWIN shows good detection accuracy and requires less energy, but still more than the remaining detectors.
The most balanced detector is HDDM\_W, which consumes little energy with very good detection accuracy.
The remaining four detectors consume even less energy, but are useless from an accuracy perspective ($<$10\%).
\end{boxB}

\subsection{RQ2: How does the type of drift influence this tradeoff?}
The visual comparison of the energy consumption between the two types of drift in Fig.~\ref{fig:energy-consumption-gradual-abrupt} indicates that three drift detectors (KSWIN, EDDM, and ADWIN) seem to consume more energy for gradual drift than for abrupt drift.
To validate these visual observations, Mann-Whitney U tests were conducted and percentage differences and Cohen's d were calculated for each drift detector.
The results are displayed in Table~\ref{tab:drift-type-energy-comparison}.

Specifically, EDDM exhibited the maximum disparity between gradual and abrupt drifts, with gradual drift consuming 8.3\% more energy.
This difference is significant with a Cohen's d of 0.77, indicating a medium to large effect.
For KSWIN, the test also resulted in a significant p-value, with gradual drift consuming 4\% more energy than abrupt drift.
This difference led to a Cohen's d of 0.32, representing a small effect size.
While ADWIN also displayed a difference, with gradual drift consuming 1.7\% more energy than abrupt drift, this test was not statistically significant.
The same is true for the remaining four detectors, where the differences between abrupt and gradual drift were between 0.2 and 0.5\%.
A further, more fine-grained analysis was conducted to analyze this difference also between the dataset generators and base classifiers.
Even for these conditions, the findings indicated that gradual drift tended to consume on average slightly more energy than abrupt drift in the specified detectors.
Nonetheless, there were also a few instances where the opposite was true, such as KSWIN detecting abrupt drift using a decision tree base classifier and the dataset generated with the mixed generator.
Detailed results are available in the replication package.

\begin{table*}[!ht]
    \centering
    \caption{Mann-Whitney U tests for energy consumption of drift types (Holm-Bonferroni adjusted p-values, $\alpha=0.05$, significant p-values in bold, ordered by effect size)}
    \begin{tabular}{lllrrrrr}
        \textbf{Detector} & \textbf{Type 1} & \textbf{Type 2} & \textbf{p-value} & \textbf{T1 Mean} & \textbf{T2 Mean} & \textbf{Difference (\%)} & \textbf{Cohen's d} \\
        \hline
        \hline
        EDDM & abrupt & gradual & \textbf{2.61E-17} & 0.0058 & 0.0064 & 8.3 & 0.77 \\
        KSWIN & abrupt & gradual & \textbf{2.30E-07} & 0.0173 & 0.0181 & 4.0 & 0.32 \\
        ADWIN & abrupt & gradual & 5.02E-01 & 0.0091 & 0.0093 & 1.7 & -- \\
        DDM & abrupt & gradual & 1.00E+00 & 0.0058 & 0.0057 & 0.5 & -- \\
        HDDM\_A & abrupt & gradual & 1.00E+00  & 0.0059 & 0.0058 & 0.4 & -- \\
        PageHinkley & abrupt & gradual & 1.00E+00 & 0.0054 & 0.0054 & 0.2 & -- \\
        HDDM\_W & abrupt & gradual & 1.00E+00 & 0.0059 & 0.0059 & 0.2 & -- \\
        \hline
        \hline
    \end{tabular}
    \label{tab:drift-type-energy-comparison}
\end{table*}

Regarding the impact of drift types on accuracy, Table~\ref{tab:accuracy-drift-detectors} already showed that most detectors reached a better true alarm rate for gradual drift.
Notable exceptions were PageHinkley and KSWIN, with the latter performing substantially better for gradual drift (93\% vs. 57\%).
However, when drift was successfully detected, the detection closeness was generally better for abrupt drift, except in the case of EDDM and HDDM\_A.

\begin{boxB}
\textbf{Answer to RQ2:}
We find that most drift detectors consume almost the same amount of energy for both types of drift, with gradual drift showing a tendency for slightly more energy consumption than abrupt drift.
Exceptions were EDDM and KSWIN, which both consume significantly more energy when detecting gradual drifts, namely 8.3\% for EDDM and 4.0\% for KSWIN.
In terms of detection accuracy, most detectors perform better with gradual drifts than abrupt ones, except for KSWIN and PageHinkley.
\end{boxB}

\subsection{RQ3: How does the type of base classifier influence this tradeoff?}

To explore this relationship, we employed the Spearman rank correlation test to evaluate whether there is a statistically significant correlation between the used base classifier and the energy consumption of the drift detector.
The outcomes of the analysis suggest that, based on the collected data, there is no discernible and statistically significant correlation between the choice of base classifier and the energy consumption of the drift detector.
Detailed information, including the comprehensive table of test results, is available in the replication package.

Additionally, to further understand the interplay between the drift detection closeness and the choice of base classifier, we also applied the Spearman rank correlation test.
The test results indicated that there is also no significant correlation between these two factors, with the minimum p-value recorded at 0.122.
This finding supports the notion that the selected base classifier does not significantly influence the drift detection closeness.
Further details can be explored in the replication package that provides a comprehensive overview of the test outcomes.

\begin{boxB}
\textbf{Answer to RQ3:}
The choice of base classifier does not have a significant impact on the energy consumption and drift detection closeness of various drift detectors.
\end{boxB}

\section{Discussion}
Based on our results, as summarzied in Fig.~\ref{fig:energy-consumption-true-alarm} and Fig.~\ref{fig:energy-consumption-closeness} we can categorize our studied detectors into three types: a) detectors that sacrifice energy efficiency for detection accuracy (KSWIN), b) balanced detectors that consume low to medium energy with good accuracy (HDDM\_W, ADWIN), and c) detectors that consume very little energy but are unusable in practice due to very poor accuracy (HDDM\_A, PageHinkley, DDM, EDDM).
From a Green AI perspective, the most compelling option for energy-efficient drift detection is HDDM\_W, which balances high accuracy with low energy consumption.
It excels in gradual drift scenarios, combining a close drift position detection with good true alarm rates.
Similar to previous work~\cite{Verdecchia2022,DelRey2023,Wei2023}, HDDM\_W is another great example that ML practitioners do not always have to choose between accuracy and energy efficiency and that there are many ML use cases where there is no or only a negligible tradeoff between these two quality attributes.
KSWIN, on the other hand, achieves the highest accuracy for abrupt drifts, but its significantly higher energy consumption makes it less attractive.
Interestingly, even for abrupt drifts, HDDM\_W offers a notable 67\% energy saving while maintaining good accuracy.
In general, the expected type of drift should slightly influence the decision for a detector, e.g., in a scenario where KSWIN is selected to be best guarded against the likely occurrence of abrupt drift, but much less than expected.
Similarly, the type of base classifier should play no role at all when choosing a detector, at least in our context of traditional ML algorithms.

While there is no study we can compare our energy efficiency results to, the results regarding detection accuracy can be compared to related work.
In another study about the accuracy of different error-based drift detection methods, it was found that ADWIN is the best performer in terms of drift position closeness and false alarms rate~\cite{Poenaru-Olaru2022}.
Our results are similar, but with the distinction that our study included all error-based drift detection methods implemented in \texttt{scikit-multiflow}, whereas the mentioned study did not include KSWIN, which was our top performer regarding accuracy in abrupt drift scenarios.
ADWIN and HDDM\_W show similar performance in accuracy metrics, but HDDM\_W is also an energy-efficient method compared to ADWIN and KSWIN.

Even though the energy consumption per detection instance is crucial in high-scalability scenarios, considering other factors is also vital.
Drift position closeness directly affects energy usage.
Earlier drift detection like KSWIN's can save energy despite higher consumption per instance, compared to less energy-intensive methods like PageHinkley with less accurate drift detection.
An increased false alarm rate may also cause more frequent model retraining, subsequently increasing energy consumption due to reduced retrain intervals.
This amount can be significant if we consider, e.g., that training OpenAI's GPT-3 a single time produces 552 tCO2 and consumes 1,287 MWh~\cite{patterson2021carbon}.
This energy can power an average U.S. household for 120 years.
From a Green AI perspective, false alarms are therefore much worse than missed alarms.

However, while false alarms can lead to unnecessary retraining, increasing energy consumption, their impact might be mitigated by optimizing retraining intervals or employing techniques to reduce false alarms.
Additionally, real-world scenarios often lack complete information like true labels and window sizes, necessitating data-distribution-based drift detection.
Investigating their energy consumption remains an open area.
Furthermore, drift width can influence performance~\cite{Poenaru-Olaru2022}.
While our study used equal lengths, exploring the impact of varying drift lengths on energy consumption would provide valuable insights.
Importantly, our findings support the conclusion of~\cite{Poenaru-Olaru2022} that error-based drift detection might not be reliable as an alarming system due to low true alarm rates.

\section{Threats to Validity}
\textbf{Internal Validity}
A potential challenge to internal validity, linked to historical factors, may have arisen in our experiment due to the potential impact of executing successive iterations on our measurements, e.g., due to rising hardware temperatures.
To address this concern, we implemented measures by introducing a 5-second sleep operation before each experimental iteration.
This ensured more uniform hardware conditions for all runs.
Likewise, a warm-up operation was conducted to guarantee that the initial iteration occurred under very similar conditions to subsequent ones, mitigating potential influences on our measurements.
As a threat to reliability of measure, the presence of background tasks during the experiment execution could have served as confounding factors, thereby affecting our energy measurements.
To address this concern, we took preemptive measures by terminating processes that were not essential for the execution of the experiment and restricted access to the infrastructure.
Furthermore, we conducted each experiment 10 times to minimize the impact of any unforeseen background processes.

\textbf{External Validity}
Regarding the generalizability of our findings, we carefully chose datasets with two primary types of concept drifts.
These datasets were generated using five synthetic dataset generators.
In addition to selecting two commonly used base classifiers, we introduced four additional base classifiers to enhance the diversity of our study.
The choice of synthetic datasets, where the drift positions are known, facilitates easy comparison of accuracy across various methods.
This deliberate selection of datasets and classifiers contributes to a more robust evaluation of the generalizability and effectiveness of the examined methods. 

\textbf{Reliability}
To ensure the reproducibility of our study, we have made the replication package accessible online.\footnote{\url{https://doi.org/10.5281/zenodo.10613150}}
Running the experiments on different hardware yielded consistent results, reinforcing the reliability of our findings and offering assurance in the robustness of the outcomes. Relying solely on a tool called CodeCarbon could threaten the construct validity of our experiment. To mitigate this risk, we took two steps. First, we leveraged its open-source nature to examine its implementation and verify its use of RAPL on Linux for gathering energy data. Second, we ran the experiment on two different systems to confirm consistent results.

\section{Conclusion}
This study investigated the tradeoff between accuracy and energy efficiency in various concept drift detectors.
The findings revealed that there is no single method that excels across all metrics.
KSWIN and ADWIN are the most energy-intensive detectors, while PageHinkley is the most energy-efficient.
On the other hand, KSWIN is the best performer in abrupt drift detection, while HDDM\_W is the best in gradual drift detection.
The type of drift influences the energy consumption of drift detectors, with gradual drift tending to consume more energy than abrupt drift for three detectors: KSWIN, EDDM, and ADWIN.
The type of drift also influences the accuracy of drift detectors, with gradual drift generally performing better in terms of true alarm rate and drift position closeness.
The base classifier does not significantly influence the energy consumption or accuracy of drift detectors.

There are several avenues for future research to build on and extend our results.

\begin{itemize}
    \item Include more drift detectors, potentially implemented in other libraries
    \item Replicate the study for various neural network architectures instead of using more traditional ML algorithms
    \item Investigate the tradeoff between accuracy and energy efficiency in online learning settings
\end{itemize}

In this study, we focused on the case of immediate label feedback, where labels are received immediately after making a prediction. 
This scenario has been the most prominent in concept drift research in recent years~\cite{Lu2018}.
Our study offers a comprehensive understanding of the accuracy vs. energy efficiency tradeoff in the methods developed for this context. 
However, in practical settings, labels are often absent or delayed. 
This reality has sparked research into novel methods for detecting model performance issues caused by concept drift, particularly through various performance estimation techniques~\cite{rabanser2019failing, sethi2017reliable, guillory2021predicting}. 
In future work, we aim to expand our research to include scenarios with delayed and absent label feedback to provide more comprehensive insights for practitioners regarding the accuracy vs. energy efficiency tradeoff in monitoring ML models.

\bibliographystyle{IEEEtranN}
\bibliography{references}

\end{document}